\documentclass[runningheads]{llncs}
\usepackage{graphicx}

\usepackage{soul}
\usepackage{xurl}
\usepackage[hidelinks]{hyperref}
\usepackage[utf8]{inputenc}
\usepackage[small]{caption}
\usepackage{graphicx}
\usepackage{amsmath,amssymb}
\usepackage{booktabs}
\usepackage{wrapfig}
\usepackage{bm}
\usepackage[linesnumbered,ruled,vlined,noend]{algorithm2e}
\usepackage{subfig}
\usepackage{color}

\newcommand{\R}{\mathbb{R}}
\newcommand{\var}{\operatorname{Var}}
\newcommand{\pageRank}{\operatorname{PR}}

\newcommand{\frane}{FRANe}
\newcommand{\numdatasets}{26}
\newcommand{\rmae}{\overline{\operatorname{RMAE}}}

\begin{document}
\title{Unsupervised Feature Ranking \\ via Attribute Networks\thanks{Supported by the Slovenian Research Agency (grant P2-0103 and a young researcher grant), and European Commission (grant 952215).}}
\author{Urh Primo\v{z}i\v{c}\inst{1}\orcidID{0000-0003-1573-7841} \and
Bla\v{z} \v{S}krlj\inst{1, 2}\orcidID{0000-0002-9916-8756} \and
Sa\v{s}o D\v{z}eroski\inst{1, 2}\orcidID{0000-0003-2363-712X} \and
Matej Petkovi\'{c}\inst{1}\orcidID{0000-0002-0495-9046}}
\authorrunning{U. Primo\v{z}i\v{c} et al.}
\institute{
Jozef Stefan Institute, Jamova 39, 1000 Ljubljana \and 
Jotef Stefan Postgraduate School, Jamova 39, 1000 Ljubljana
\\
\email{urh.primozic@student.fmf.uni-lj.si, \{blaz.skrlj,saso.dzeroski,matej.petkovic\}@ijs.si}
}
\maketitle              %
\begin{abstract}
The need for learning from unlabeled data is increasing in contemporary machine learning. Methods for unsupervised feature ranking, which identify the most important features in such data are thus gaining attention, and so are their applications in studying high throughput biological experiments or user bases for recommender systems. We propose FRANe (Feature Ranking via Attribute Networks), an unsupervised algorithm capable of finding key features in given unlabeled data set. FRANe is based on ideas from network reconstruction and network analysis. FRANe performs better than state-of-the-art competitors, as we empirically demonstrate on a large collection of benchmarks.
Moreover, we provide the time complexity analysis of FRANe further demonstrating its scalability.
Finally, FRANe offers as the result the interpretable relational structures used to derive the feature importances.
\end{abstract}

\section{Introduction}
\label{sec:introduction}

Increasing amounts of high-dimensional data, in fields such as molecular and systems' biology, require development of fast and scalable feature ranking algorithms ~\cite{vir:stanczykJainPregled}. By being able to prioritize the feature space with respect to a given target, feature ranking algorithms already offer, e.g., novel biomarker candidates.
However, the amount of available labeled data is potentially much smaller when compared to the amount of unlabeled data, which remains largely unexploited. In response, \emph{unsupervised} feature ranking algorithms (that operate only on unlabeled data) are actively developed. %

\begin{figure}[t!]
    \centering
    \includegraphics[width = .6\linewidth]{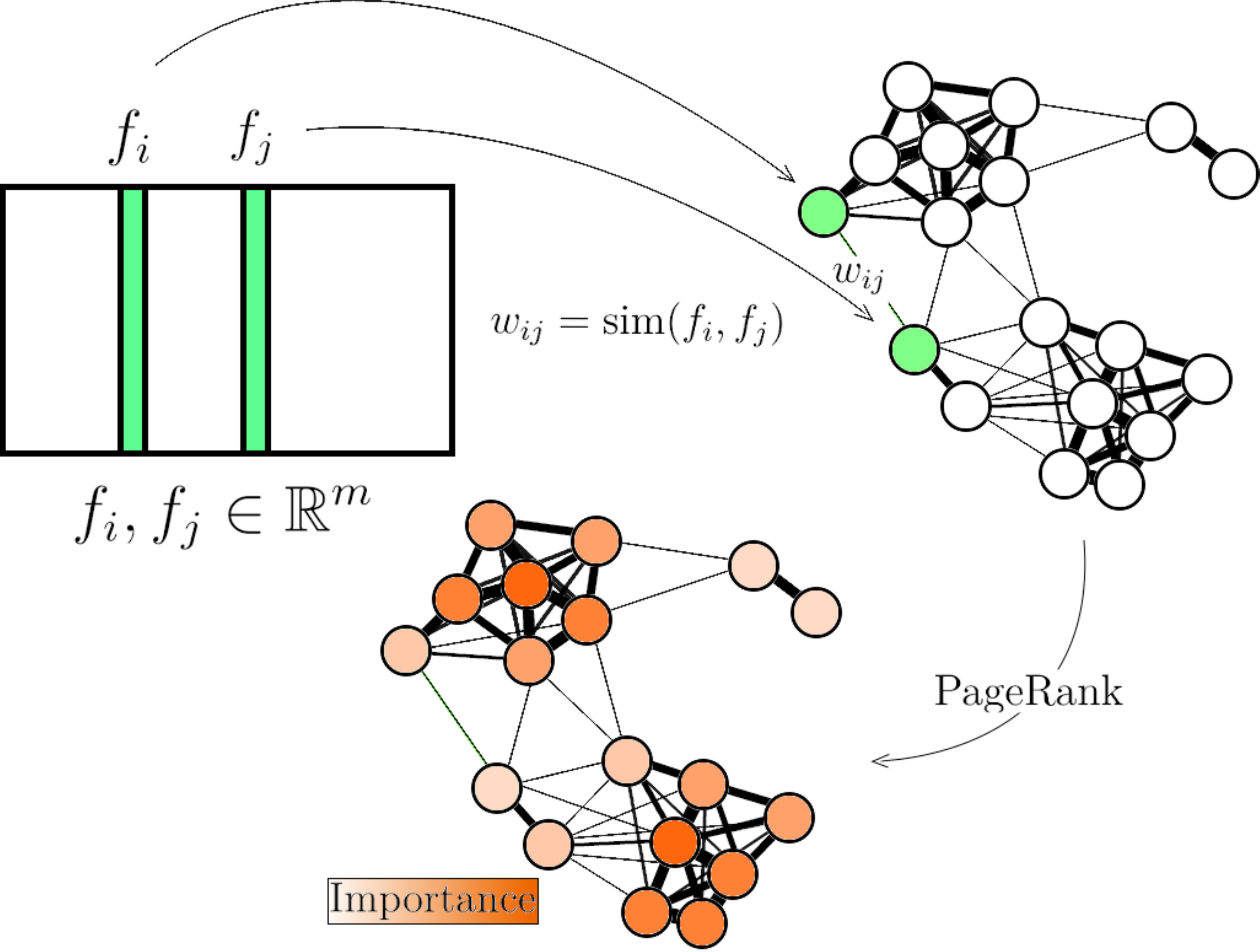}
    \caption{Overview of the FRANe method.} %
    \label{fig:shema}
\end{figure}

We propose FRANe, a {\bf F}eature {\bf R}anking approach based on {\bf A}ttribute {\bf Ne}tworks), schematically shown in Fig.~\ref{fig:shema}. FRANe achieves state-of-the-art performance by exploiting data-derived relations between the features (which form an undirected weighted graph). The contributions of this work are manifold, and can be summarized as follows:
\begin{enumerate}
    \item We propose FRANe, a fast algorithm for unsupervised feature ranking based on reconstructing attribute networks and subsequent node ranking.
    \item We demonstrate the algorithm's state-of-the-art performance on \numdatasets{} datasets, validating our claims via Bayesian and classical performance analysis.
    \item We present an extensive theoretical analysis of the proposed algorithm.
    \item We offer an implementation of FRANe as a simple-to-use, freely available Python library, which also includes other baseline approaches.
\end{enumerate}
The remainder of this work is structured as follows. In Section~\ref{sec:related}, we discuss the related work that has led us to propose FRANe. We describe the proposed method in Section~\ref{sec:method}. Next, we discuss the experimental setup (Section~\ref{sec:setup}), followed by our results (Section~\ref{sec:results}) and conclusions (Section~\ref{sec:conclusions}).

\section{Related Work}
\label{sec:related}

Unsupervised feature ranking is a relatively new research endeavor. An overview of unsupervised ranking algorithms~\cite{review:fsel} was published only recently. Some of the currently well-established methods for unsupervised feature ranking include: {\bf \textsc{Laplace}}~\cite{laplace}, %
{\bf \textsc{MCFS}}, %
and {\bf \textsc{NDFS}}. %
All of them construct a network of instances by employing an instance similarity measure. Finally, recent work -- awarded the best paper award at ECML PKDD 2019 -- uses autoencoder~\cite{agnos}: the {\bf AgnoS-S} algorithm gives feature ranking scores as a parameter vector at the early stages of a neural network, which learns to reconstruct the input space and assigns each input variable a score as a side-result.

Apart from the unsupervised feature ranking literature, we also draw inspiration from the literature on \emph{network reconstruction} and its applications in gene expression analysis~\cite{sanguinetti2019gene,Langfelder2008wgcna,chiquet2019variational}.
Network reconstruction derives a network from a tabular data set, so that relations between instances (rows) or features (columns) are identified, maintained, and used for a given down-stream task. Once a tabular data set is converted to a network (graph), various centrality measures can be used to determine the centrality of the nodes in the network. Our method uses PageRank centrality measure \cite{pagerank} and its generalization to weighted graphs. While we use it in the unsupervised fashion (somewhat similarly to \cite{8919274}), it can be also used in the supervised scenario \cite{weightedPageRank}.

\section{Method}
\label{sec:method}
Real data often consists of groups of similar features. Intuitively, each such group has a \textbf{representative feature} that is most similar to all others. This feature can be expected to predict the values of the other features in the group reasonably well, making the others \textbf{redundant}. Thus, the most central features are potentially good candidates for a set of features that a feature selection algorithm would return.
When the number of features in the data goes into thousands and more, it is expected that many of them are effectively random noise or completely redundant. The corresponding noisy weights could prevent discovering the wanted centrality values: We therefore introduce a \textbf{minimal weight threshold} and only connect the features that are similar enough.

It is not clear in advance which threshold value is the best. Therefore, we try out a set of candidate thresholds, following geometric threshold progression and ranging from the minimal to the maximal edge weight. We calculate the centrality (feature importance) values from the corresponding graphs, and obtain a set of feature rankings.
Among those, we choose the one that maximizes the heuristic that is based on the intuition that the feature importance values in a good ranking have a large spread.
Rankings obtained with low thresholds are expected to be similar, whereas small increases of high thresholds can cause large changes in the rankings. Sets of candidate thresholds with few low thresholds and many large ones, e.g., the geometric threshold sequence, are expected to give good results.

\subsection{Algorithm}

Let $X = [x_{i, j}]_{i, j}\in\R^{m\times n}$ be a data set, where $m$ is the number of examples and $n$ is the number of features. The $i$-th example (row in the matrix $X$), $1\leq i \leq m$, is given as $\bm{x}_i = [x_{i, 1}, \dots, x_{i, n}]$. The $j$-th feature, $1\leq j\leq n$, is given as a feature vector (column in the matrix $X$) $\bm{f}_j = [x_{1, j}, \dots, x_{m, j}]^T$.

The computation of \frane{} is given in Alg.~\ref{alg:frane}. At input, it takes the (training) data $X$, a minimal edge threshold and the number of iterations $I$. First, it computes the feature similarly matrix $W=[w_{j, k}]\in\mathbb{R}^{n\times n}$.
It then computes the geometric sequence $\bm{T}$ of (edge-weight) thresholds as follows.
First, we define the set of similarities between different features $W' = \{w_{j,k} \mid j \neq k \}$,
together with $M' = \max(W')$ and $m' = \min(W')$. Then, the dissimilarity values
$D = \{M' - w | w\in W' \land w < M'\}$ are computed. Finally, the thresholds $t_i\in \bm{T}$ are defined as
$\bm{T}=[t_1, \ldots , t_{I}]$ where
\begin{equation}
    \label{eqn:thresholds}
    t_i = M' - \text{min}(D) \cdot \left ( \frac{\text{max}(D)}{\text{min}(D)} \right )^{(i - 1) / (I - 1)} 
\end{equation}
The temporary resort to dissimilarities is necessary, because we want to analyze the region of larger similarities more thoroughly.
For every threshold $t_i \in \bm T$, we build a weighted graph $G(t_i)$ with $n$ vertices that correspond to features. An edge with the weight $w_{j,k}$ between $\bm f_j$ and $\bm f_k$ exists in $G(t_i)$, if $w_{j, k} \geq t_i$.
To avoid too sparse graphs, we consider only those, for which the average degree $\bar{e} = |\{w_{j, k} | j < k \land w_{j, k} \geq t_i \}| / n$ exceeds $\bar{e}_\text{min} = 1$.

We run a PageRank on the graph $G(t_i)$, which returns a possible ranking $\bm{r}(t_i) = [\text{PR}(\bm{f}_1), \ldots , \text{PR}(\bm{f}_n)]$, where $\pageRank(\bm{f}_j)$ is the PageRank importance $\pageRank{}(j)$ of the node of feature $\bm f_j$ in $G(t_i)$,
as defined in \cite{weightedPageRank,pagerank}.
After iterating through all thresholds, calculating the rankings $\bm r(t_i)$ for each $t_i \in \bm T$, we pick as output the ranking with the highest value of the ranking quality heuristic RQH, where
\begin{equation}
    \label{eqn:hrq}
    \text{RQH}(\bm{r}) = \frac{\text{second largest score in } \bm{r}}{\text{second smallest score in } \bm{r}}\text{.} 
\end{equation}
The second largest and smallest scores are taken for stability reasons as the medians of the three largest and smallest scores, respectively.
\begin{algorithm}
\DontPrintSemicolon %
\caption{\frane{}($X$, $\bar{e}_\text{min}$, $I$)}\label{alg:frane}
    $W =$ compute
 $[w_{j, k}]_{j, k = 1}^n = [\mathit{PearsonCorr}(\bm f_j, \bm f_k) + 1]_{j, k = 1}^n$ \tcp*{$w_{jk}\geq 0$}
    $S = []$\hfill \tcp*{candidate rankings}
    $\bm{T} = $ list of $I$ thresholds $t_i$ \tcp*{Eq.~\eqref{eqn:thresholds}}
    \For{$t_i \in \bm{T}$}
    {   
        $\bar{e} = |\{w_{j, k} | j, k \land w_{j, k} \geq t_i \}| / n$  \tcp*{Avoid sparse graphs}
        \If{$\bar{e} \geq \bar{e}_\text{min}$}{
            $\bm r$ = PageRank($G(t_i)$) \\ %
            add $\bm{r}$ to $S$\\ %
        }
    }
    \Return $\operatorname{arg max}_{\bm{r}\in S} \operatorname{RQH}(\bm{r})$ \tcp*{Eq.~\eqref{eqn:hrq}}%
\end{algorithm}

The first step of the algorithm requires the computation of pairwise similarities, yielding time complexity of $\mathcal{O}(m n^2)$.
Then, all the graphs $G(t_i)$ can be constructed in the total time of $\mathcal{O}(n^2)$, if we start with a fully connected graph and then incrementally remove the edges with the weights on the intervals $[t_{i - 1}, t_{i})$.
Using the power method for PageRank and assuming that the number of iterations is upper-bounded with some constant \cite{pagerank}, computing PageRank takes $\mathcal{O}(n^2)$ steps. Thus, the total number of steps in the algorithm is $\mathcal{O}(m \cdot n^2 + I \cdot n^2)$. Note that the most time-consuming step (similarity computation) can be easily parallelized, and that computing PageRank demands only vectorizable matrix-vector multiplication and vector-vector addition. 

\section{Experimental Setup}\label{sec:setup}
In this section, we describe the experimental procedure that we employ to investigate the following questions: i) How does \frane{} compare to state-of-the-art methods for unsupervised feature ranking, and ii) What is the influence of the different parameters or \frane{} on its performance?

We first give a brief description of the data sets used, continue with the evaluation procedure and finish with the parametrization of the methods.
Note that the code that allows for \textbf{replicating our experiments} (including the computation of training and testing splits) is freely available at
\url{https://github.com/FRANe-team/FRANe-dev}.

We obtained the data from the Scikit-feature repository \cite{scikit-feature}.
We wanted to use all the datasets, but had to exclude three data sets from the study (\texttt{orlraws10P}, \texttt{lung-small} and \texttt{warpAR10P}) to meet the independence assumptions of the statistical tests.
Table \ref{tab:data} gives a more detailed description of the data, including their domains.
\begin{table}[h]
  \centering
  \caption{Number of features ($n$), examples ($m$) and the domain of the used benchmarks.}
  \label{tab:data}
    \begin{tabular}{|l|rrl|c|l|rrl|}
\cline{1-4}\cline{6-9}
            & $n$ & $m$ & domain  && & $n$ & $m$ & domain   \\
\cline{1-4}\cline{6-9}
gli-85      &22283&85&  biology     && glioma      &4434&50&  biology       \\      	
smk-can-187 &19993&187&  biology            && relathe     &4322&1427&  text data          \\
cll-sub-111 &11340&111&  biology            && lymphoma    &4026&96&  biology             \\
arcene      &10000&200&  mass spectrometry  && lung        &3312&203&  biology            \\
pixraw10p   &10000&100&  face image         && pcmac       &3289&1943&  text data          \\
nci9        &9712&60&  biology             && warppie10p  &2420&210&  face image         \\
carcinom    &9182&174&  biology            &&colon       &2000&62&  biology             \\
allaml      &7129&72&  biology             &&coil20      &1024&1440&  face image         \\
leukemia    &7070&72&  biology             && orl        &1024&400&  face image          \\
prostate-ge &5966&102&  biology            && yale        &1024&165&  face image         \\
tox-171     &5748&171&  biology            && isolet      &617&1560&  speech recognition \\
gisette     &5000&7000&  digit recognition  && madelon     &500&2600&  artificial         \\
baseshock    &4862&1993&  text data          && usps        &256&9298&  drawings \\
\cline{1-4}\cline{6-9}
\end{tabular}
\end{table}
When evaluating the feature ranking algorithms, we follow the approach of \cite{agnos}. Here, an algorithm is evaluated via 10-fold cross-validation. For a given partition of a data set into test part (one of the folds) and train part (the remaining 9 folds), feature ranking is computed on the train part.
Then, the $n'$ top-ranked features are selected and the $5$ nearest neighbor ($5$NN) model that uses only these features for predicting the values of all the features is trained (on the train part of the data).
Finally, the performance of the feature ranking algorithm is measured in terms of the predictive performance of the $5$NN model on the test set.
As evaluation measure, we use the average relative mean absolute error
$
\rmae{} =
\frac{1}{n}\sum_{i = 1}^n
    \frac{1}{m_\text{TEST}}
        \sum_{j = 1}^{m_\text{TEST}}
            \frac{\left| \hat{x}_{ij} - x_{ij} \right|}{\sigma(\bm{f}_i)}
$,
where $m_\text{TEST}$ is the number of examples in the test set,
$\hat{x}_{ij}$ is the $5$NN's prediction for $x_{ij}$, and $\sigma(\bm{f}_i) = \sqrt{\var{}(\bm{f}_i)}$
is the standard deviation of the feature $\bm{f}_i$. A low value of $\rmae{}$ means that the subset of $n'$ chosen features can well reconstruct all the feature values.

The obtained $\rmae{}$ values are averaged over the $10$ folds. To see how the predictive performance of $5$NN changes as more and more top-ranked features are considered, one can build a series of $5$NN models that use $n' \in \{1, 2, \dots, 2^k \} \cup \{ n \}$, where $2^k \leq n < 2^{k+1}$ features, as shown in Fig.~\ref{fig:many}.
This may be more informative, but is harder to analyze when comparing different algorithms through statistical tests. For such comparisons, performance at $n' = 16$ is chosen. 
The hierarchical Bayesian t-test considered in this work is discussed in more detail in~\cite{bayesianTests}. The test approximates the posterior probability of the difference in performance between a pair of classifiers. The posterior plot can be visualized as a simplex, where each point represents a sample from the posterior distribution. By counting such samples in different parts of the simplex, the probability of one classifier outperforming the other is estimated.

The number of iterations in \frane{} was set to $I = 100$ and the threshold for the average number of edges was set to $\bar{e}_\text{min} = 1$. For the decay factor $\delta$ in PageRank, the recommended value of $\delta = 0.85$ was used. For other algorithms, we used the recommended parameter values. Additionally, the number of clusters for the methods MCFS and NDFST was set to the number of classes in the datasets at hand. This was possible since we used classification datasets from the Scikit-feature repository. The classes were otherwise ignored.

\section{Results}\label{sec:results}
In this section, we first report the results of the comparison between \frane{} and its competitors. We then focus on different parts of \frane{} and consider alternative design choices. 

The $\rmae{}$ values for the different feature ranking methods, i.e., the corresponding $5$NN models, are shown in Table \ref{tab:results}.
\begin{table} [!htb]
  \centering
  \caption{The performance (measured in terms of $\rmae{}$) of $5$NN models that use the $n'= 16$ top-ranked features from a given feature ranking. The last two rows of the table additionally give the average rank of each algorithm and its number of wins, i.e., the number of times it is ranked first. The best result in each row is shown in bold.}
  \vspace{0.3cm}
  \hspace*{-2mm}~\resizebox{.8\textwidth}{!}{
    \begin{tabular}{lrrrrrr}
    \hline
          & \multicolumn{1}{l}{FRANe} & \multicolumn{1}{l}{Laplace} & \multicolumn{1}{l}{NDFS} & \multicolumn{1}{l}{Agnos-S} & \multicolumn{1}{l}{MCFS} & \multicolumn{1}{l}{SPEC} \\
    \hline
gli-85      & 0.745 &  \textbf{0.736} & 0.775 & 0.774 & 0.747 &  0.797 \\ 
smk-can-187 & 0.610 & 0.612 & 0.62 & 0.656 & 0.626 &  \textbf{0.597} \\
cll-sub-111 &  \textbf{0.716} & 0.738 & 0.736 & 0.763 & 0.77 &  0.777 \\ 
arcene      & 0.759 &  \textbf{0.457} &  \textbf{0.457} & 0.734 &  \textbf{0.457} &  0.733 \\ 
pixraw10p   &  \textbf{0.348} & 0.412 & 0.412 & 0.352 & 0.412 &  0.377 \\ 
nci9        &  \textbf{0.763} & 0.771 & 0.771 & 0.839 & 0.771 &  0.807 \\ 
carcinom    & 0.719 & 0.739 & 0.751 & 0.743 &  \textbf{0.717} &  0.743 \\ 
allaml      &  \textbf{0.711} & 0.726 & 0.747 & 0.775 & 0.744 &  0.749 \\ 
leukemia    &  \textbf{0.824} & 0.833 & 0.833 & 0.857 & 0.833 &  0.836 \\ 
prostate-ge & 0.485 & 0.503 &  \textbf{0.482} & 0.552 & 0.509 &  0.649 \\ 
tox-171     &  \textbf{0.725} & 0.77 & 0.785 & 0.734 & 0.776 &  0.781 \\ 
gisette     &  \textbf{0.440} & 0.481 & 0.481 & 0.509 & 0.481 &  0.533 \\ 
baseshock    & 0.174 & 0.188 &  \textbf{0.163} & 0.182 & 0.191 &  0.197 \\ 
glioma      &  \textbf{0.609} & 0.643 & 0.636 & 0.716 & 0.615 &  0.685 \\ 
relathe     & 0.182 &  \textbf{0.174} & 0.183 & 0.284 & 0.187 &  0.218 \\ 
lymphoma    &  \textbf{0.774} & 0.873 & 0.873 & 0.804 & 0.873 &  0.873 \\ 
lung        &  \textbf{0.700} & 0.708 & 0.734 & 0.749 & 0.701 &  0.780 \\ 
pcmac       & 0.156 &  \textbf{0.147} & 0.160 & 0.168 &  \textbf{0.147} &  0.163 \\ 
warppie10p  & 0.370 & 0.526 & 0.526 &  \textbf{0.316} & 0.526 &  0.526 \\ 
colon       &  \textbf{0.652} & 0.661 & 0.661 & 0.666 & 0.661 &  0.661 \\ 
coil20      & 0.234 & 0.364 &  \textbf{0.205} & 0.407 & 0.786 &  0.528 \\ 
orl        & 0.572 & 0.703 & 0.703 &  \textbf{0.479} & 0.703 &  0.703 \\ 
yale        & 0.608 & 0.749 & 0.749 &  \textbf{0.572} & 0.749 &  0.749 \\
isolet      & 0.567 & 0.548 & 0.562 &  \textbf{0.523} & 0.619 &  0.643 \\ 
madelon     &  \textbf{0.853} & 0.856 & 0.856 & 0.86 & 0.856 &  0.856 \\ 
usps        & 0.371 & 0.338 &  \textbf{0.283} & 0.337 & 0.422 &  0.394 \\ 
\hline
average rank & \textbf{1.88} & 2.54 & 2.88 & 4.04 & 3.19 & 4.42 \\
number of wins & \textbf{12} & 4 & 5 & 4 & 3 & 1 \\
\hline
    \end{tabular}%
    }
  \label{tab:results}%
\end{table}%
We can see that \frane{} outperforms its competitors. First of all, it has the best average rank (1.88) among the considered algorithms.
The second best algorithm (in terms of the average rank) is Laplace with an average rank of 2.54.
The difference between \frane{} and the other algorithms is even more visible when we compare the numbers of wins: \frane{} is the best performing algorithm in 12 cases (46\% win rate). The second highest number of wins (5) is achieved by NDFS.

To also show some statistical evidence for the quality of the \frane{} rankings, we employ the Bayesian hierarchical t-test \cite{bayesianTests}, since it directly answers which of the two compared algorithms is better.
The other popular option -- frequentist non-parametric tests such as Friedman and Bonferroni-Dunn \cite{demsar} -- allow for comparison of more than one algorithm, but these tests are typically too weak (as follows from their definitions \cite{demsar}), and are harder to interpret.

\begin{figure}[!htb]
\centering
    \centering
    \includegraphics[width=0.7\linewidth]{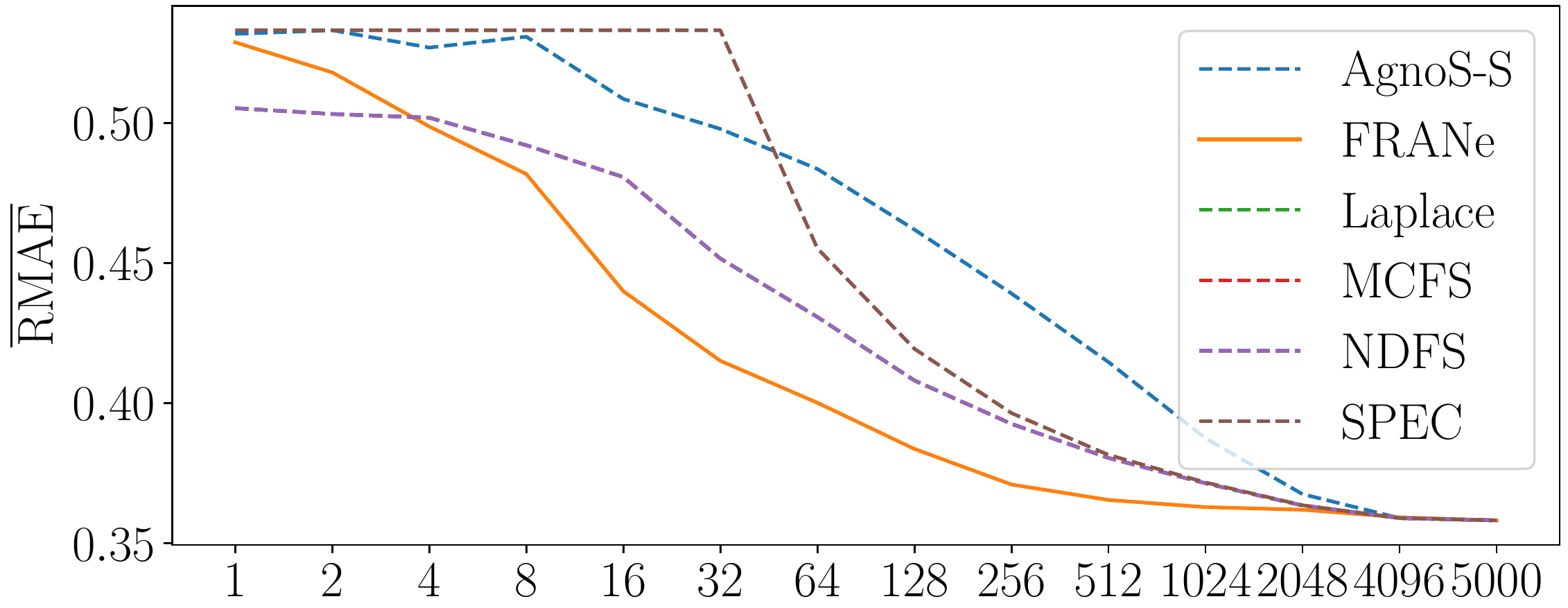}%
\caption{Error curves for the different rankings on Gisette dataset.}
\label{fig:many}
\end{figure}
\noindent The Bayesian comparison indicates that \frane{} dominates its closest competitor (Laplace), in $26\%$ of the cases, whereas the Laplace method is better in only $2\%$ of the cases. In the other cases, the difference in performance is smaller than 0.001 and is considered practically insignificant. This is consistent with the results in Table~\ref{tab:results}: the overall win-rate of \frane{} is notably higher (12 against 4), even though these two algorithms differ by less than one in average rank values.
A detailed (and more global) comparison of the rankings (where the number of chosen features varies from $1$ to $n$) on Gisette data set is given in Fig.~\ref{fig:many}.
It is clear that the \frane{} rankings are the best as its corresponding curve is below the curves of all other rankings. 

\subsection{Alternative Design Choices}
After we have proved that \frane{} offers state-of-the-art performance, we now investigate the sensitivity of its performance to varying its key components. Due to space constraints,
we only vary the similarity measure used in the computation of the matrix $W$, the threshold progression that defines the list of edge-weight thresholds $\bm{T}$,
and the ranking quality heuristic RQH, while the node-centrality measure is left fixed (PageRank), and left for further work. We first give a brief description of the considered threshold progressions and similarity measures between \emph{different} features $W' = \{w_{j,k} \mid j \neq k  \}$, with $m' = \min W'$, and $M' = \max W'$.

\begin{paragraph}{Similarity measures.}
Let $\bm f_j = [x_{1,j}, \ldots , x_{m,j}], \bm f_k =[x_{1,k}, \ldots , x_{m,k}] \in \mathbb R^m$ be two feature vectors.
Besides correlation, other similarity measures can be used. They are all based on different distance measures $d(\bm f_j, \bm f_k)$: i) \textbf{Canberra} ($\sum_{i=1}^m \frac{|x_{i,k} - x_{i, j}|}{|x_{i, j}| + |x_{i, k}|}$, ii) \textbf{Chebyshev}  ($\max_{i=1}^m |x_{i,k} - x_{i, j}|$), iii) \textbf{Manhattan} ($\sum_{i=1}^m |x_{i,k} - x_{i, j}|$), and iv) \textbf{Euclidean} ($\left(\sum_{i=1}^m |x_{i,k} - x_{i, j}|^2\right)^{1/2}$).
The corresponding similarity measures are defined as
$\operatorname{sim}(\bm f_j \bm f_k) = M' - d(\bm f_j \bm f_k)$.
\end{paragraph}

\begin{paragraph}{Threshold functions.}
The definition of the thresholds $t_i$ from Eq. \eqref{eqn:thresholds} originally follows the \textbf{geometric} progression. The alternatives are:
i) \textbf{Linear}($m', M'$),
ii) \textbf{Linear}($\operatorname{mean}(W'), M'$),
iii) \textbf{Linear}($\operatorname{median}(W'), M'$),
and iv) \textbf{Quantile}, where $t_i = i$-th $I$-quantile of $W'$'s for the latter,
and $t_i = \frac{b - a}{I - 1}(i - 1) + a$ for $\textbf{Linear}(a, b)$. The motivation for using linear progression that starts at the mean (or its more stable analogue the median) of the $W'$ values is that, intuitively, larger thresholds are more interesting to analyze, since the corresponding graphs are sparser.
\end{paragraph}

\begin{figure}
    \centering
    \includegraphics[width=0.5\linewidth]{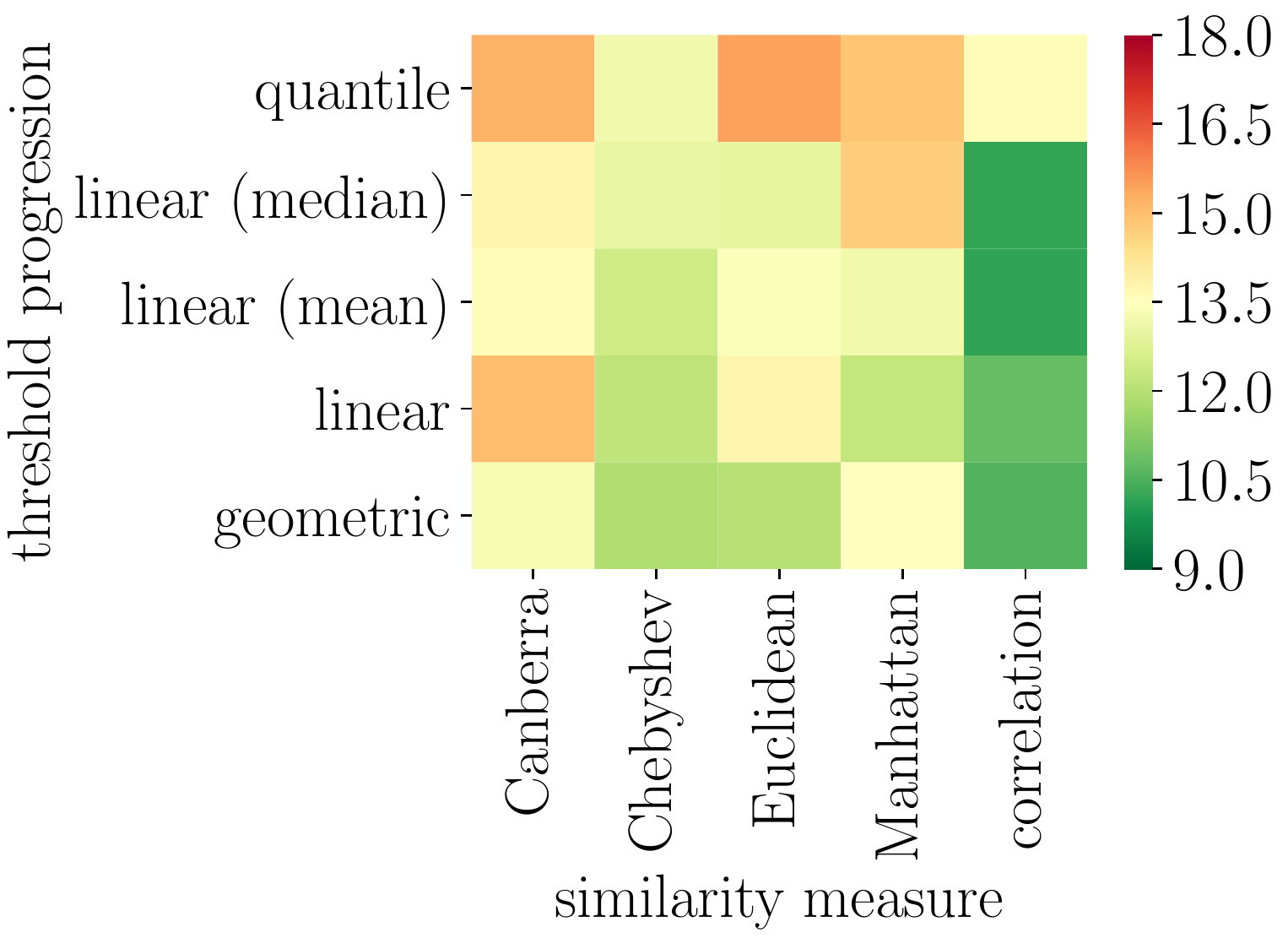}
    \caption{Average ranks (over datasets) of different combinations similarity metric - threshold progression. The legend denotes the average rank of a given metric-progression combination (the lower, the better).}
    \label{fig:ablation}
\end{figure}

The results (see Fig.~\ref{fig:ablation}) show that
\frane{} is quite robust with respect to the chosen threshold progression and to the chosen similarity measure. Except for the correlation similarity (works best for 10/26 data sets), and the geometric threshold progression (works best in 9/26 cases), all the similarity measures and threshold progressions perform approximately equally well. Still, no fixed (progression, similarity) pair has more than 3 wins. The detailed results are available at \url{https://github.com/FRANe-team/FRANe}). They also include the experiments with RQH, where we show that RQH outperforms random search (in 22/26 cases), which is often considered a strong baseline in optimization \cite{randomSearch}. 

\section{Conclusion}
\label{sec:conclusions}
In this work we have presented FRANe, an algorithm for network-based unsupervised feature ranking. In contrast to existing approaches, FRANe attempts to \emph{reconstruct} a representative network of \emph{features}. By ranking \emph{nodes} in this network via the efficient PageRank approach, we achieve state-of-the-art results for the task of unsupervised feature ranking.

The results indicate that the proposed unsupervised ranking algorithm is indeed a strong competitor to the existing approaches. Theoretical analysis indicates the $\mathcal{O}(n^2)$ complexity of the distance computation as one of the main bottlenecks. The current implementation of FRANe, however, exploits highly optimized compiled routines and scales seamlessly for each of the considered data sets. An extension which would reduce the quadratic complexity could include random subspace sampling (where the probability of choosing a feature depends on its variance). 

The proposed methodology is suitable from the interpretability point of view, as the key nodes (features) and their, e.g., correlation-based neighborhoods are easily inspected. This can  potentially offer novel insights into key parts of the feature space governing a given data set's structure.

Given that the main spatial bottleneck is related directly to computation of PageRank scores (maintaining the graph in the memory), we believe that an option for further scalability could potentially include distributed storage-based matrix operations  \cite{pageRankHadoop,pageRankDistr2}, which would facilitate ranking of attributes when considering very large data sets.

As further work, we believe that distances between features could be also computed in latent space, where embeddings of features would be first obtained (via the transposed feature matrix), potentially speeding up the correlation computation, as well as providing more robust rankings. Furthermore, the body of work related to \emph{metric learning} could similarly prove useful when determining the most suitable similarity score.

\bibliographystyle{splncs04}
\bibliography{ijcai21}

\end{document}